\documentclass{article}
\usepackage{iclr2020_conference,times}
\usepackage{listings}
\usepackage{graphicx}
\usepackage{hyphenat}
\usepackage{adjustbox}
\setcitestyle{square,sort,comma,numbers}
\usepackage{hyperref}
\usepackage{enumerate}
\usepackage{enumitem}

\title{\protect\raggedright{A Reinforcement Learning Environment for Mathematical Reasoning via Program Synthesis}}

\author{Joseph Palermo \& Johnny Ye \\
	\textbf{Cash App Labs} \\
	\texttt{\{jpalermo, jye\}@squareup.com} \\
	\AND
	\textbf{Alok Singh} \\
	\textbf{Lawrence Berkeley National Laboratory} \\
	\texttt{alok@lbl.gov} \\
}

\graphicspath{ {./} }
\iclrfinalcopy

\begin{document}

\maketitle

\begin{abstract}
	We convert the DeepMind Mathematics Dataset into a reinforcement learning environment by interpreting it as a program synthesis problem. Each action taken in the environment adds an operator or an input into a discrete compute graph. Graphs which compute correct answers yield positive reward, enabling the optimization of a policy to construct compute graphs conditioned on problem statements. Baseline models are trained using Double DQN on various subsets of problem types, demonstrating the capability to learn to correctly construct graphs despite the challenges of combinatorial explosion and noisy rewards.
\end{abstract}

\section{Introduction}

The DeepMind Mathematics Dataset \cite{DeepmindMath} consists of synthetically generated math problems. They cover a range of problem types including: Numbers, comparison, measurement, arithmetic, algebra, polynomials, calculus, and probability. These problem types are arranged into a collection of 56 modules each containing different sub-types of problems. The dataset provides the problems in the form of question-answer pairs represented as ASCII text.

The predominant algorithmic approach for learning to produce the answers conditioned on the question statements has been to train seq2seq models \cite{DeepmindMath}\cite{TPTransformer}\cite{ScalingLaws}. For some of the modules this approach yields very nearly 100\% accuracy, however for other modules this does extremely poorly. For example, with this approach the best reported test accuracy for the module "numbers\_\_list\_prime\_factors" is less than 25\% \cite{ScalingLaws}.

This poor performance is not surprising considering the nature of the problems on which this occurs. For instance, the module "numbers\_\_is\_prime" requires discriminating primes from non-primes. Apart from memorizing answers, there are only a few useful hacks that a learned model can easily pick up, for example, the fact that any even number (except for 2) is not prime. Algorithms to correctly test for primality require a sequence of divisions to be performed very precisely, and before a perfectly correct algorithm is obtained loss may not be much improved beyond random guessing. Thus, it's unclear if the gradient of loss obtained by comparing computed answers to correct answers could be sufficient to learn such algorithms.

A further difficulty with such systems is that they are unlikely to be sufficiently accurate or interpretable for real-world use. Models trained in this manner are effectively black boxes which only return answers. Probability estimates assigned to tokens sampled from the model can potentially provide clues about the likelihood of correctness (assuming they are well calibrated). However, the reasoning process by which the result was arrived at cannot be easily inspected.

We think it would be desirable to train neural networks to make use of existing programs for computing various mathematical operations \cite{BertACalculator}\cite{NeuralModuleNetworks}. Just as human programmers use libraries when writing larger programs, learned algorithms should be able to build upon existing operators. With predefined operators available, the neural network wouldn't have to rediscover for itself algorithms and functions which are already well known, and could instead focus on learning how to compose them. Essentially, we propose to treat the problem from the point of view of program synthesis.

Program synthesis is the problem of how to automatically construct programs to meet predefined specifications or constraints \cite{ProgSynth}. If the specification is sufficiently precise, then one may be able to prove that a given program meets it. However, the DeepMind Mathematics Dataset provides only question-answer pairs. So, if a program was generated to compute the answer for a given question, in general it would only be possible to confirm that the program generated the correct answer for that specific question.

More formally, the learning problem is to learn a policy which maps a problem statement represented as text, to a program. The program takes the form of a compute graph composed of discrete operators. A policy which constructs these compute graphs by conditioning on problem statements can be learned with reinforcement learning because the question-answer pairs can be used to provide a reward signal to adjust the parameters of the policy.

\section{Framing the Problem using Reinforcement Learning}

To frame the problem in terms of reinforcement learning we need to define an environment. To explain how this is done we first need to explain how inputs are extracted from problem statements and to specify how the operators are defined.

\subsection{Inputs}

In problems in the DeepMind Mathematics Dataset there are often parts of the text which are explicitly mathematical. We call these the inputs of the problems.

For example, in the following problem:

\begin{lstlisting}
Let h(t) = t**3 + t**2 + 1. Let v(d) = 6*d**3 + 24*d**2 + 4. Let w(j) = 4*h(j) - v(j). What is the third derivative of w(x) wrt x?
\end{lstlisting}

The inputs are:

\begin{lstlisting}
1) h(t) = t**3 + t**2 + 1
2) v(d) = 6*d**3 + 24*d**2 + 4
3) w(j) = 4*h(j) - v(j)
4) w(x)
5) x
\end{lstlisting}

It turns out that for many problem types, a parser can be written which automatically extracts the inputs of the problem. By embedding such a parser within the environment implementation, given any problem statement the environment can determine which inputs are available. This will be used to help define the action space in section 2.3.

\subsection{Predefined Operators}

To utilize program synthesis, a list of operators needs to be defined. Each of them takes some fixed number of inputs and returns a single output. For example, one operator called 'differentiate\_wrt' (short for “differentiate with respect to” takes as input an expression and a variable, and it outputs the derivative of the expression with respect to the variable.

A full list of operators is defined in Appendix A.

\subsection{Defining the Reinforcement Learning Environment}

Observations from the environment consist of a representation of the question concatenated with a sequence of indices corresponding to actions taken previously in the current episode.

The representation of the question can either be provided in encoded form (i.e. as an array of indices into a vocabulary of tokens) or unencoded form (i.e. raw text) depending on how the environment is configured. If the environment is configured to return questions in encoded form, then it uses a predefined byte pair encoding constructed from a corpus of questions. If the representation is provided in encoded form then it is also padded up to a fixed maximum length. Note that the raw text of the question is available at every step regardless of how configurations are set (through the info object returned by the "step" method of the environment).

The action space is discrete and corresponds to the set of operators and available inputs which can be introduced into the graph at any given time. The graph is built up in breadth first order. Fixing the order of graph construction simplifies the action space because otherwise the actions would need to both specify which operator is being applied and where in the graph it is being introduced.

The maximum number of available inputs is set as a parameter of the environment. Thus if the max number of available inputs is set to n\_inputs, and there are n\_ops operators, then the action space is represented by the set of integers from 0 to n\_ops + n\_inputs - 1. The inputs are represented within the action space in the order in which they appear in the problem. For example, given the following problem:

\begin{lstlisting}
Let h(t) = t**3 + t**2 + 1. Let v(d) = 6*d**3 + 24*d**2 + 4. Let w(j) = 4*h(j) - v(j). What is the third derivative of w(x) wrt x?
\end{lstlisting}

The action space would be as follows:

\begin{lstlisting}
0 : 1st operator
1 : 2nd operator
...
n_ops - 1 : nth operator
n_ops + 0 : h(t) = t**3 + t**2 + 1
n_ops + 1 : v(d) = 6*d**3 + 24*d**2 + 4
n_ops + 2 : w(j) = 4*h(j) - v(j)
n_ops + 3 : w(x)
n_ops + 4 : x
n_ops + 5 : None
...
n_ops + n_inputs - 1 : None
\end{lstlisting}

Note that if less than the maximum number of possible inputs is present in a given problem, then the actions after the last available input are simply defined as None and are masked (see section 3.1.1).

The reward has a value of 1 if the compute graph computes the right answer and a value of 0 otherwise. Note that if the compute graph is incomplete then it computes the value None which always yields a reward of 0.

Here is an example trajectory from the environment in which the environment state is provided in unencoded format for legibility. Note that 5 is the action index corresponding to the "differentiate" operator and 14 is the action index corresponding to the 1st input (which in this case corresponds to 6*k**2 - 101*k + 2548):

\begin{tabular}{rl}
\textbf{state  t=0 :} & What is the first derivative of 6*k**2 - 101*k + 2548?; \\
\textbf{action t=0 :} & 5 \\
\textbf{state  t=1 :} & What is the first derivative of 6*k**2 - 101*k + 2548?; 5 \\
\textbf{reward t=1 :} & 0 \\
\textbf{action t=1 :} & 14 \\
\textbf{state  t=2 :} & What is the first derivative of 6*k**2 - 101*k + 2548?; 5, 14 \\
\textbf{reward t=2 :} & 1 \\ 
\end{tabular}

Note that the environment will automatically terminate episodes with reward 0 if the configured maximum length is reached.

\subsection{Modules}

For the purposes of this research, we restricted consideration to a subset of modules from the DeepMind Mathematics Dataset that have composed counterparts. The reason for this is that the composed problems require much larger graphs to reliably compute correct answers. This means that in a reinforcement learning context, these problems require longer action sequences to deliver optimal reward, and so constitute a more challenging reinforcement learning environment.

Furthermore, it is the case that some of the modules without uncomposed counterparts contain problems for which some of the relevant inputs are represented simply as words and so are difficult to parse as inputs. For example the module 'measurement\_\_conversion' contains questions such as: "What is seven halves of a day in minutes?" 

A full list of supported modules is defined in Appendix A.

\section{Challenges}

\subsection{Combinatorial Explosion}

The experiments we present here use 15 operators and require up to 3 inputs, which gives an action space of size 18. We also set a limit on graph size of at most 7 nodes. Even with this restriction, there are \begin{math} 18^7 \approx 6.12 * 10^8\end{math} possible graphs. Given such a large state-space, naive search would be unlikely to stumble upon correct graphs. Here we present strategies for mitigating the impact of combinatorial explosion.

\subsubsection{Masking Invalid Actions}

To reduce the effective size of the search space we mask actions which are guaranteed to produce invalid graphs. In particular we introduce a type hierarchy and assign types to the parameter(s) and output of each operator. Also, each input is automatically assigned a type by the environment. Taken together, typed operators and typed inputs allow the application of type constraints limit which actions are valid at any given time. To implement this we define a method of the environment (called compute\_mask) which  generates a boolean masking vector over the action space. This method can be called by an agent before selecting an action to determine which actions are invalid, although formally the environment does not force this constraint. So invalid actions can still be taken, but they will result in graph that fails to compute a meaningful result (i.e. output of the graph will be "None").

Recall that actions add nodes to the graph in breadth-first order. Thus, each  action after the first will correspond to a specific parameter of an operator that has already been applied. The type of the parameter being filled-in by the action determines the type constraint applied to that action. The form of the constraint is that if the parameter requires type X, then only types at or below X in the type hierarchy are permitted.

The type hierarchy consists of the following custom types: Equation (e.g. 2*x + y = 3), Expression (e.g. 2*x + 1), Function (e.g. f(x) = 2*x + 1), Value (e.g. 2), Variable (e.g. x), Rational (e.g. 1/2). The hierarchy for the custom types is provided in Figure 1. We also utilize the following types which are built into Python: object, list, dict, and bool. Note that list, dict, bool and all the custom types are subclasses of object, and thus fall below it in the type hierarchy \cite{Python}.

\begin{figure}
  \centering
  \includegraphics[scale=0.7]{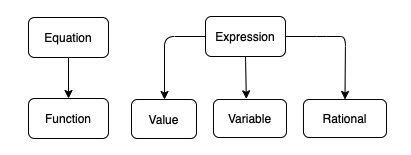}
  \caption{Type Hierarchy}
\end{figure}

In addition to type constraints, there are two other additional constraints implemented by masking. The first action always corresponds to the root node of the graph and has the additional constraint that it must correspond to an operator (i.e. it cannot be an input). There is also the additional constraint that any action defined as None is always masked. This occurs for problems which have less than the maximum number of inputs in which case the action space is padded up to a fixed size with actions defined as None.
 
\subsubsection{Abstracting Subgraphs into Operators} 

Another strategy for addressing combinatorial explosion which can be applied but was not implemented in our experiments is the process of abstracting frequently rewarded subgraphs into new operators. The concept is to identify subgraphs which frequently occur in graphs that yield reward, then to identify the inputs and output of that subgraph such that it could be redefined as a new operator. This is known more broadly in the computer science literature as Frequent Subgraph Mining \cite{FSM}.

For example:

\begin{lstlisting}
differentiate_wrt(differentiate_wrt(Expression('-3*z**5 + 13*z**3 + 41*z**2'),Variable('z')),Variable('z')) reward = 1
\end{lstlisting}

This subgraph can be abstracted into a new operator as follows:

\begin{lstlisting}
diff_wrt_2(p0, p1) = differentiate_wrt(differentiate_wrt(p0, p1), p1)
\end{lstlisting}

New operators defined in this way could be introduced into the action space and would provide the possibility of finding shorter graphs to compute the same functions. This is similar to the operator abstraction process in Dreamcoder \cite{Dreamcoder}. This would increase the size of the action space and hence the number of possible graphs, however it would reduce the required length of frequently rewarded graphs and hence would have a multiplicative effect in reducing the effective size of the search space.

\subsection{Noisy Reward}

An additional challenge with this environment is that the rewards are noisy with respect to graph correctness. In other words, a particular graph might compute the right answer for a given question but do so in the wrong way. For example:

\begin{lstlisting}
Is 5340 a multiple of 10?; not(is_prime(Value('10'))) = True, reward: 1
\end{lstlisting}

If the inputs of the problem were changed the same graph may no longer compute the right answer because it is simply performing the wrong operations and so cannot reliably generalize to different inputs. This effectively means that a positive reward from the environment does not guarantee that the constructed graph is correct.

\section{Approach}

We use Double DQN \cite{DDQN} to learn a value function that maps a state and an action to the expected discounted sum of rewards. An off-policy learning algorithm is used because it simplifies the maintenance of exploration which is critical in the presence of noisy rewards. Epsilon-greedy exploration is utilized with a step-wise linear annealing schedule on the value of epsilon.

We also apply prioritized experience replay \cite{PrioritizedReplay} to improve the efficiency of learning. We sample steps from replay memory with priority directly proportional to the most recently computed TD-error. After every training batch, we re-compute replay priorities for both the steps used to construct the batch and an additional random sample of steps from the replay buffer. The additional random sample is taken to encourage all replay priorities to remain consistent with the current model parameters, even if they haven't been used to construct a training batch in some time.

We initialize the replay buffer with trajectories (containing up to 50k steps) collected by a uniform random policy, where a one-to-one balance is kept between trajectories with positive reward and trajectories with 0 reward. This balancing is done to prevent a heavy skew towards trajectories with reward 0, due to reward sparsity in the environment. After an initial period of training on the replay buffer as initialized, new experience is continuously incorporated into the replay buffer. However, a one-to-one balance between trajectories with positive reward and 0 reward is maintained since otherwise trajectories with 0 reward would overwhelm the replay buffer.

The model used in experiments is a transformer encoder \cite{Transformer} with 6 encoder blocks followed by 2 dense layers. We use 4 attention heads and a hidden layer size of 256. The dense blocks use a hidden layer size of 256. We apply dropout \cite{Dropout} of 0.1 in the Transformer blocks and between the dense layers. We use a learning rate of 5*10e-5 and a batch size of 512. The epsilon value for epsilon-greedy exploration is initialized to 0.4 and is linearly annealed to 0.05 by an increment of 2.5*10e-5 per step. The full set of hyperparameters used to produce the results reported here is provided in the code (\href{https://github.com/joepalermo/dm\_math\_solvers/blob/master/hparams-for-paper.cfg}{https://github.com/joepalermo/dm\_math\_solvers/blob/master/hparams-for-paper.cfg}).

\raggedbottom

\section{Results}

We conduct experiments on different subsets of modules to evaluate how it affects model performance. For each experiment we run 5 trials with the same hyperparameters and different random seeds (as recommended in Henderson et al. 2017 \cite{RLThatMatters}). We sample 1.01 million examples from each module under consideration with 800k/200k/10k across train/validation/test. The trials are run for 50k steps each except for the trials in the "Interference" experiment (see below) which are run for 100k steps.

In Table 1 we report mean test reward for the median trial. We define the median trial as the trial for which the mean test reward across modules is the median across trials. In appendix B we report results from all trials and also provide validation curves corresponding to the results reported in Table 1.

In our experiments we consider only the uncomposed modules. Composed modules are distinguished by containing much longer problems (they are the modules suffixed by "\_composed" or "\_compose"). Based on preliminary experiments, the methods employed here do not perform well on the composed modules due to the increased challenge of combinatorial explosion. Note also that in the experiments we report here the "calculus\_differentiate" module is filtered to remove multivariate problems as this reduces the maximum required graph size (the option to remove this filter is provided as a hyperparameter of the environment).

In these experiments we also limit the action space to contain only the 15 operators required to successfully compute correct graphs on the modules selected. The list of operators used is provided in appendix B.2.

In our first experiment we train on all uncomposed modules simultaneously and observe significant interference between modules reflected by lower final test performance than in smaller subsets of modules. We hypothesize that by including multiple modules in which answers to problems are similarly expressed, interference between modules is magnified. For example, the modules "numbers\_\_is\_factor" and "numbers\_\_is\_prime" both have true/false answers. In the case in which both modules are trained on simultaneously the operators "divides" and "is\_prime" respectively will be frequently misused during exploration (e.g. Is 5340 a multiple of 10? not(is\_prime(Value('10'))) = True, which results in a positive reward).

To investigate this hypothesis we select two additional subsets of modules. The first of these module subsets contains "numbers\_\_is\_factor" and "numbers\_\_is\_prime", and we refer to it as the "Inteference" experiment. The second contains "numbers\_\_is\_prime", "numbers\_\_list\_prime\_factors", "numbers\_\_calculus\_differentiate", "numbers\_\_div\_remainder", and "numbers\_\_gcd", and we refer to it as the "No Interference" experiment. The modules in the "Interference" experiment are selected because the form of the answers in those modules are both true/false and so will result in the type of collisions described above (i.e. noisy rewards). The modules in the "No Interference" experiment are selected because the form of the answers in those modules is such that they are unlikely to result in the type of collisions we describe above. However, notably the final results show that "numbers\_\_is\_prime" had better performance in the "Interference" experiment which is evidence against the hypothesis.

\begin{center}
\begin{tabular}{ | | c | c | c | c | |}
\hline
\multicolumn{4}{|c|}{Test Results} \\
\hline
 Module & Interference& No Interference & All Uncomposed Modules \\ 
 \hline
 \hline
 numbers\_\_is\_factor & 0.7800 & - & 0.3669 \\  
 \hline
 numbers\_\_is\_prime & 1.000 & 0.7382 & 0.6567 \\   
  \hline
 numbers\_\_list\_prime\_factors & - & 1.000 & 1.000 \\   
  \hline
 calculus\_\_differentiate & - & 0.8511 & 0.3350 \\   
  \hline
 polynomials\_\_evaluate & - & - & 0.9517 \\   
  \hline 
 numbers\_\_div\_remainder & - & 1.000 & 0.8159 \\   
  \hline
 numbers\_\_gcd & - & 1.000 & 0.9990 \\   
  \hline
 numbers\_\_lcm & - & - & 1.000 \\
  \hline
 algebra\_\_linear\_1d & - & - & 0.1124 \\   
  \hline
 algebra\_\_polynomial\_roots & - & - & 0.8253 \\
 \hline
 algebra\_\_linear\_2d & - & - & 0.2660 \\
 \hline
 \hline
 Mean Reward across Modules & 0.8900 & 0.9179 & 0.6830 \\ 

\hline
\end{tabular}
\label{tab:test-results}Table 1: Test reward per module for the median trial of each experiment.
\end{center}

\section{Discussion}

There are two main contributions of this paper. We have introduced a new reinforcement learning environment by interpreting the DeepMind Mathematics Dataset as a problem in program synthesis. We have also trained a baseline model on several subsets of the uncomposed modules despite the dual challenge of combinatorial explosion and noisy rewards. 

From preliminary experiments it's clear that performance on significantly longer graphs is poor, however we believe that by implementing subgraph abstraction (as described in section 3.1.2) performance on longer graphs could be significantly improved. Furthermore we suspect that content based attention (as in \cite{ContentBasedAttention}) could be a useful architectural component to integrate into the learned model due to the dynamic nature of the action space. We consider these to be interesting topics for future research.

\section*{Code}

We provide a light-weight repository containing an implementation of the reinforcement learning environment and setup instructions:

\href{https://github.com/JohnnyYeeee/math\_prog\_synth\_env}{https://github.com/JohnnyYeeee/math\_prog\_synth\_env}

\vspace{8pt}

We also provide the full code used to produce the results reported here:

\href{https://github.com/joepalermo/dm\_math\_solvers}{https://github.com/joepalermo/dm\_math\_solvers}

\section*{Acknowledgments} 

The authors would like to thank Alex Krizhevsky, Rayhane Mama, Hashiam Kadhim, Marc Tyndel, and Ragavan Thurairatnam for helpful discussions.

\newpage

\newpage
\appendix
\section{Environment}

\subsection{Predefined Operators}

We provide here a list of the predefined operators. The numbers associated to them indicates their index in the action space. They are also provided along with a type signature (e.g. f(x: type\_of\_x) -$>$ return\_type).

\begin{enumerate}
\setcounter{enumi}{-1}

\item \begin{lstlisting} 
lookup_value(mapping: Dict[Variable: Value], key: Variable) -> object
\end{lstlisting}
Given a dictionary and a key to query it, it returns the corresponding value.

\item \begin{lstlisting} 
solve_system(system: List[Equation]) -> Dict[Variable: Value]
\end{lstlisting}
Given a system of equations returns a dictionary mapping variables to values.

\item \begin{lstlisting} 
append(system: List[Equation], equation: Equation) -> List[Equation]
\end{lstlisting}
Given a list of equations, appends a new equation to the end of that list.

\item \begin{lstlisting} 
append_to_empty_list(equation: Equation) -> List[Equation]
\end{lstlisting}
Returns a new list containing only the given equation.

\item \begin{lstlisting} 
factor(inpt: Expression) -> Expression
\end{lstlisting}
Converts a polynomial into irreducible factors over rational numbers.

\item \begin{lstlisting} 
differentiate(expression: Expression) -> Expression
\end{lstlisting}
Returns the first derivative of a polynomial. (This function assumes univariate)

\item \begin{lstlisting} 
mod(numerator: Value, denominator: Value) -> Value
\end{lstlisting}
Returns the remainder of the numerator divided by the denominator

\item \begin{lstlisting} 
gcd(x: Value, y: Value) -> Value
\end{lstlisting}
Returns the greatest common divisor of x and y

\item \begin{lstlisting} 
divides(numerator: Value, denominator: Value) -> bool
\end{lstlisting}
Returns True if denominator is divisible by numerator

\item \begin{lstlisting} 
is_prime(x: Value) -> bool
\end{lstlisting}
Returns True if x is prime and False otherwise

\item \begin{lstlisting} 
lcm(x: Value, y: Value) -> Value
\end{lstlisting}
Returns the least common multiple of x and y

\item \begin{lstlisting} 
lcd(x: Rational, y: Rational) -> Value
\end{lstlisting}
Given two rationals return the least common denominator

\item \begin{lstlisting} 
prime_factors(n: Value) -> Set[Value]
\end{lstlisting}
Returns a set of all prime factors of n

\item \begin{lstlisting} 
evaluate_function(function_definition: Function, function_argument: Expression) -> Value
\end{lstlisting}
Evaluates a function by substituting the variable in function\_definition by the function\_argument. function\_argument can either look like '2' or 'f(2)'

\item \begin{lstlisting}
not_op(x: bool) -> bool
\end{lstlisting}
Returns the inverse of a boolean.

\item \begin{lstlisting} 
differentiate_wrt(expression: Expression, variable: Variable) -> Expression
\end{lstlisting}
Returns the first derivative of an expression with respect to a given variable.

\item \begin{lstlisting} 
make_equation(expression1: Expression, expression2: Expression) -> Equation
\end{lstlisting}
Returns an equation where expression1 is set equal to expression2.

\item \begin{lstlisting}
simplify(inpt: object) -> object
\end{lstlisting}
Returns a simplification of inpt based on sympy heuristics.

\item \begin{lstlisting}
make_function(expression1: Expression, expression2: Expression) -> Function
\end{lstlisting}
Returns a function where expression1 is set to be equal to expression2.

\item \begin{lstlisting}
replace_arg(function: Function, var: Variable) -> Function
\end{lstlisting}
Replaces the argument in function with the given variable.

\item \begin{lstlisting}
lookup_value_equation(mapping: Dict[Variable: Value], key: Variable) -> Equation
\end{lstlisting}
Given a dictionary and a key to query it, it returns an equation of "key = value"

\item \begin{lstlisting}
extract_isolated_variable(equation: Equation) -> Variable
\end{lstlisting}
Given an equation it returns the isolated variable.

\item \begin{lstlisting}
substitution_left_to_right(arb: object, eq: Equation) -> object
\end{lstlisting}
Returns the arb with all found instances of the equation's left hand side substituted by the equation's right hand side.

\end{enumerate}

\subsection{Supported Modules}

Here is the full list of supported modules:

\begin{lstlisting}
numbers__is_factor
numbers__is_prime
numbers__list_prime_factors
calculus__differentiate
polynomials__evaluate
numbers__div_remainder
numbers__gcd
numbers__lcm
algebra__linear_1d
algebra__polynomial_roots
algebra__linear_2d
algebra__linear_1d_composed
algebra__linear_2d_composed
algebra__polynomial_roots_composed
calculus__differentiate_composed
numbers__div_remainder_composed
numbers__gcd_composed
numbers__is_factor_composed
numbers__is_prime_composed
numbers__lcm_composed
numbers__list_prime_factors_composed
polynomials__evaluate_composed
polynomials__compose
\end{lstlisting}

\section{Experiments}

\subsection{Full Test Results}

\begin{adjustbox}{max width=\textwidth,center}
\begin{tabular}{| | c | c | c | c | c | c | |}
\hline
\multicolumn{6}{|c|}{Full Test Results for All Uncomposed Modules} \\
\hline
Module & Run 1  & Run 2  & Run 3  &Run 4& Run 5 \\
\hline \hline 
numbers\_\_is\_factor           & 0.2907       & 0.5239       & 0.2798       & 0.4807       & 0.3669      \\ \hline  
numbers\_\_is\_prime            & 0.7294       & 1.000            & 0.4860        & 1.000  & 0.6567      \\ \hline 
numbers\_\_list\_prime\_factors & 1.000       & 1.000            & 1.000            & 1.000  & 1.000           \\ \hline
calculus\_\_differentiate       & 0.3238       & 0.3611       & 0.6095  & 0.3377      & 0.3350       \\ \hline
polynomials\_\_evaluate         & 0.9990        & 0.9950        & 0.0000            & 0.9942       & 0.9517      \\ \hline
numbers\_\_div\_remainder       & 1.000            & 1.000            & 0.7944      & 1.000            & 1.000           \\ \hline
numbers\_\_gcd                  & 0.9981       & 0.8132       & 0.9317       & 1.000            & 0.9990       \\ \hline
numbers\_\_lcm                  & 1.000            & 1.000            & 1.000            & 1.000 & 1.000           \\ \hline
algebra\_\_linear\_1d           & 0.7470       & 0.7470       & 1.000 & 1.000            & 0.1124      \\ \hline
algebra\_\_polynomial\_roots    & 0.4443       & 0.7159       & 0.9457       & 0.9143       & 0.8253      \\ \hline
algebra\_\_linear\_2d           & 0.8110        & 0.1400        & 0.9600         & 1.000            & 0.2660       \\ \hline
\hline
Mean reward across modules      & 0.6973 & 0.6816 & 0.6629 & 0.8843 & 0.6830 \\
\hline
\end{tabular}
\end{adjustbox}
\begin{center}
\label{tab:test-results}Table 2: Test reward per module for runs on all uncomposed modules
\end{center}

\begin{adjustbox}{max width=\textwidth,center}
\begin{tabular}{| | c | c | c | c | c | c | |}
\hline
\multicolumn{6}{|c|}{Full Test Results for "Interference" Experiment} \\
\hline
Module & Run 1  & Run 2  & Run 3  &Run 4& Run 5 \\
\hline \hline 
numbers\_\_is\_factor           & 0.5155       & 0.7893       & 0.9383       & 0.9922       & 0.7800      \\ \hline  
numbers\_\_is\_prime            & 0.5229       & 0.9980            & 0.5258        & 1.000  & 1.000      \\ \hline 
\hline
Mean reward across modules      & 0.5192 & 0.8937 & 0.7321 & 0.9961 & 0.8900 \\
\hline
\end{tabular}
\end{adjustbox}
\begin{center}
\label{tab:test-results}Table 3: Test reward per module for "Interference" experiment
\end{center}

\begin{adjustbox}{max width=\textwidth,center}
\begin{tabular}{| | c | c | c | c | c | c | |}
\hline
\multicolumn{6}{|c|}{Full Test Results for "No Interference" Experiment} \\
\hline
Module & Run 1  & Run 2  & Run 3  &Run 4& Run 5 \\
\hline \hline 
numbers\_\_is\_prime            & 0.5563       & 0.5400            & 0.7694        & 1.000  & 0.7382      \\ \hline 
numbers\_\_list\_prime\_factors & 1.000       & 1.000            & 1.000            & 1.000  & 1.000           \\ \hline
calculus\_\_differentiate       & 0.8112       & 0.9572       & 1.000  & 0.9653      & 0.8511       \\ \hline
numbers\_\_div\_remainder       & 1.000            & 1.000            & 1.000  & 1.000  & 1.000           \\ \hline
numbers\_\_gcd                  & 0.9990       & 1.000            & 1.000  & 1.000  & 1.000       \\ \hline
\hline
Mean reward across modules      & 0.8733 & 0.8994 & 0.9539 & 0.9931 & 0.9179 \\
\hline
\end{tabular}
\end{adjustbox}
\begin{center}
\label{tab:test-results}Table 4: Test reward per module for "No Interference" experiment
\end{center}

\subsection{Operators Selected for Experiments}
\begin{enumerate}
    \item lookup\_value
    \item solve\_system
    \item append
    \item append\_to\_empty\_list
    \item factor
    \item differentiate
    \item mod
    \item gcd
    \item divides
    \item is\_prime
    \item lcm
    \item lcd
    \item prime\_factors
    \item evaluate\_function
    \item not\_op
\end{enumerate}

\subsection{Validation Curves}
In the below graphs the dark centre line shows the median of the five trials and the shaded area bounds the 10th and 90th percentiles based on linear interpolation.
\begin{figure}[!htb]
  \includegraphics[scale=1]{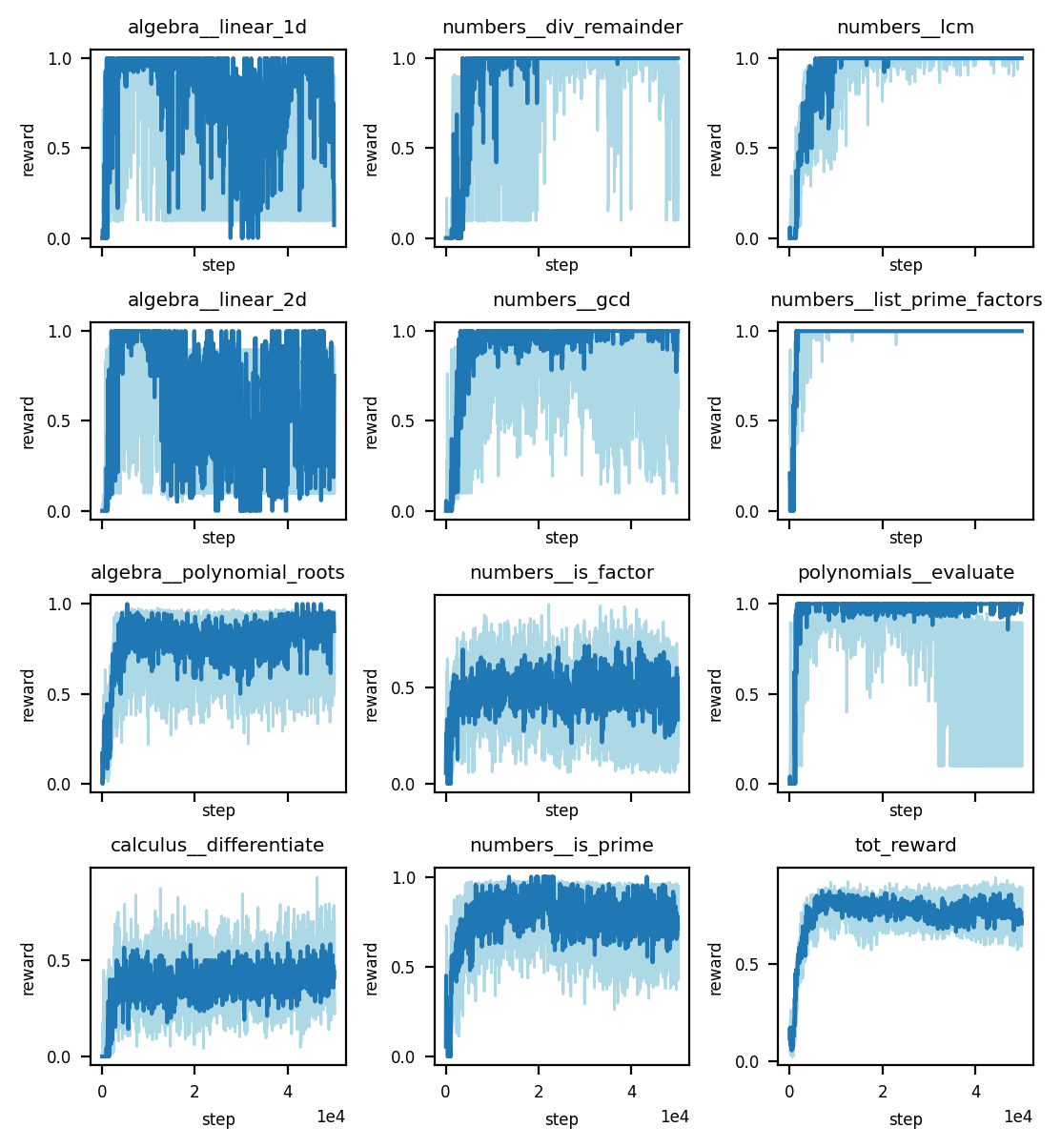}
  \caption{Reward for all uncomposed modules over 50000 steps.}
\end{figure}

\begin{figure}[!htb]
  \includegraphics[scale=1]{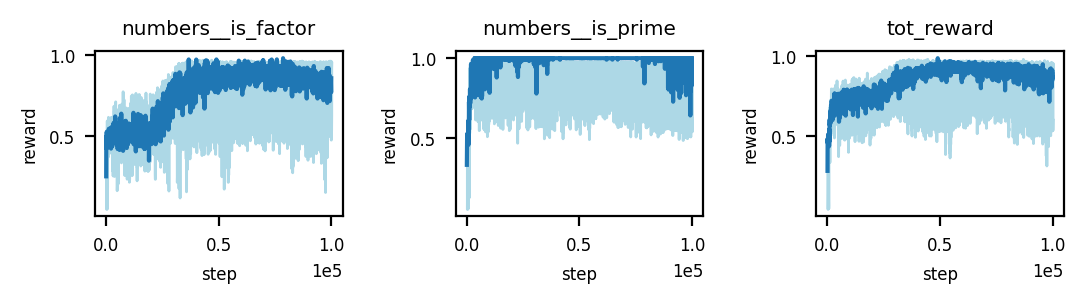}
  \caption{Reward for "Interference" experiment over 100000 steps}
\end{figure}

\begin{figure}[!htb]
  \includegraphics[scale=1]{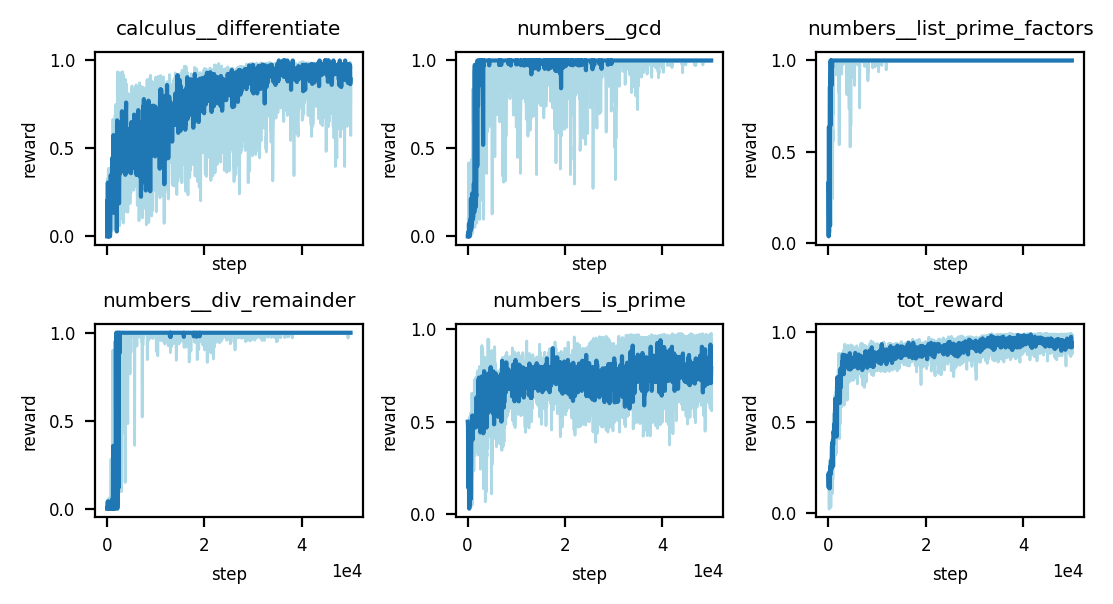}
  \caption{Reward for "No Interference" experiment over 50000 steps.}
\end{figure}

\end{document}